# Deep Reinforcement Learning in Surgical Robotics: Enhancing the Automation Level

Cheng Qian# and Hongliang Ren*

#Department of Electrical Engineering and Information Technology, Technical University of Munich, Germany

*Department of Electronic Engineering, The Chinese University of Hong Kong, Hong Kong

Abstract:

Surgical robotics is a rapidly evolving field that is transforming the landscape of surgeries. Surgical robots have been shown to enhance precision, minimize invasiveness, and alleviate surgeon fatigue. One promising area of research in surgical robotics is the use of reinforcement learning to enhance the automation level. Reinforcement learning is a type of machine learning that involves training an agent to make decisions based on rewards. This literature review aims to comprehensively analyze existing research on reinforcement learning in surgical robotics. The review identified various applications of reinforcement learning in surgical robotics, including pre-operative, intra-body, and percutaneous procedures, listed the typical studies, and compared their methodologies and results. The findings show that reinforcement learning has great potential to improve the autonomy of surgical robots. Reinforcement learning can teach robots to perform complex surgical tasks, such as suturing and tissue manipulation. It can also improve the accuracy and precision of surgical robots, making them more effective at performing surgeries.

## I. Introduction

The use of surgical robots has significantly increased in the last decade, driven by the need for precision, safety, and efficiency in surgeries [1]. Since the appearance of da Vinci robotic-assisted surgical system in 2000 [3], surgical robots have proven to help perform minimally invasive surgeries (MIS), providing better visualization, higher precision, and reduced invasiveness, and helping reduce surgeons' fatigue [2]. However, the full potential of surgical robots has yet to be realized, and there is still a need to improve their autonomy. In the last decade, more and more studies have been conducted on autonomous surgical robots [4]. To achieve autonomy in surgery, it is crucial for robots to understand the surgical task objectives, perceive complex physical environments, and autonomously make decisions. One of the biggest challenges in the autonomy of surgical robots is the high variance of surgical tasks [13], which is hard to address by explicitly modeling and planning. Therefore, Artificial Intelligence (AI) solutions emerged due to the model-free property and learning capability.

Deep Reinforcement Learning (DRL), a deep learning-based planning method that allows robots to learn from interaction with environments in a semi-supervised fashion without a pre-defined model, is one of the most promising approaches [5]. DRL has been increasingly highlighted in recent years, since its success in Atari [7]. It has demonstrated the possibility of enabling intelligent agents to outperform human experts in multiple fields. Compared to conventional planning methods, DRL provides advantages, e.g., end-to-end learning, complex decision-making, generalization, and transferability, handling uncertainties, and continuous learning. These properties enable DRL to handle high-dimensional inputs from cameras and sensors in surgeries, apply its acquired knowledge and skills to different patients, handle unforeseen variations, and continuously learn and refine their performance during surgery procedures. For these reasons, in the context of surgical robots, DRL provides a powerful model-free framework and a set of tools for learning various complicated surgical tasks with complex physical environments, which are hard to model [6]. Many studies have utilized DRL on robots under abundant surgical scenarios, e.g., ultrasound scanning, cutting and sewing, tissue retraction, needle steering, and catheterization. Currently, there have been some reviews on DRL in the scope of medical imaging, e.g., radiation therapy, image registration [8], health care application, e.g., clinical decision support [9, 10, 12], medicine, e.g., medicine treatment or development [11]. However, there still lacks a review specifically on the applications of DRL in medical robots. To fill this gap, this chapter will present a literature review highlighting the typical state-of-art works in the past 5 years (2018-2023) that utilize DRL in autonomous surgical robots, and comparing their methodologies, limitations, and results. We divide these works into three categories according to their access modes, namely:

1) Extra-body skin-interfaced procedure
2) Intra-body procedure
3) Percutaneous procedure,

which are three main procedures that we find the combination of DRL and surgical robot is mainly applied to. The various surgical tasks learned by the robot in this review are illustrated in Fig. 1, including steerable needle planning in keyhole neurosurgery, needle insertion in ophthalmic microsurgery, neck vessel and spine US scanning, tissue cutting and retraction, and wound suturing.

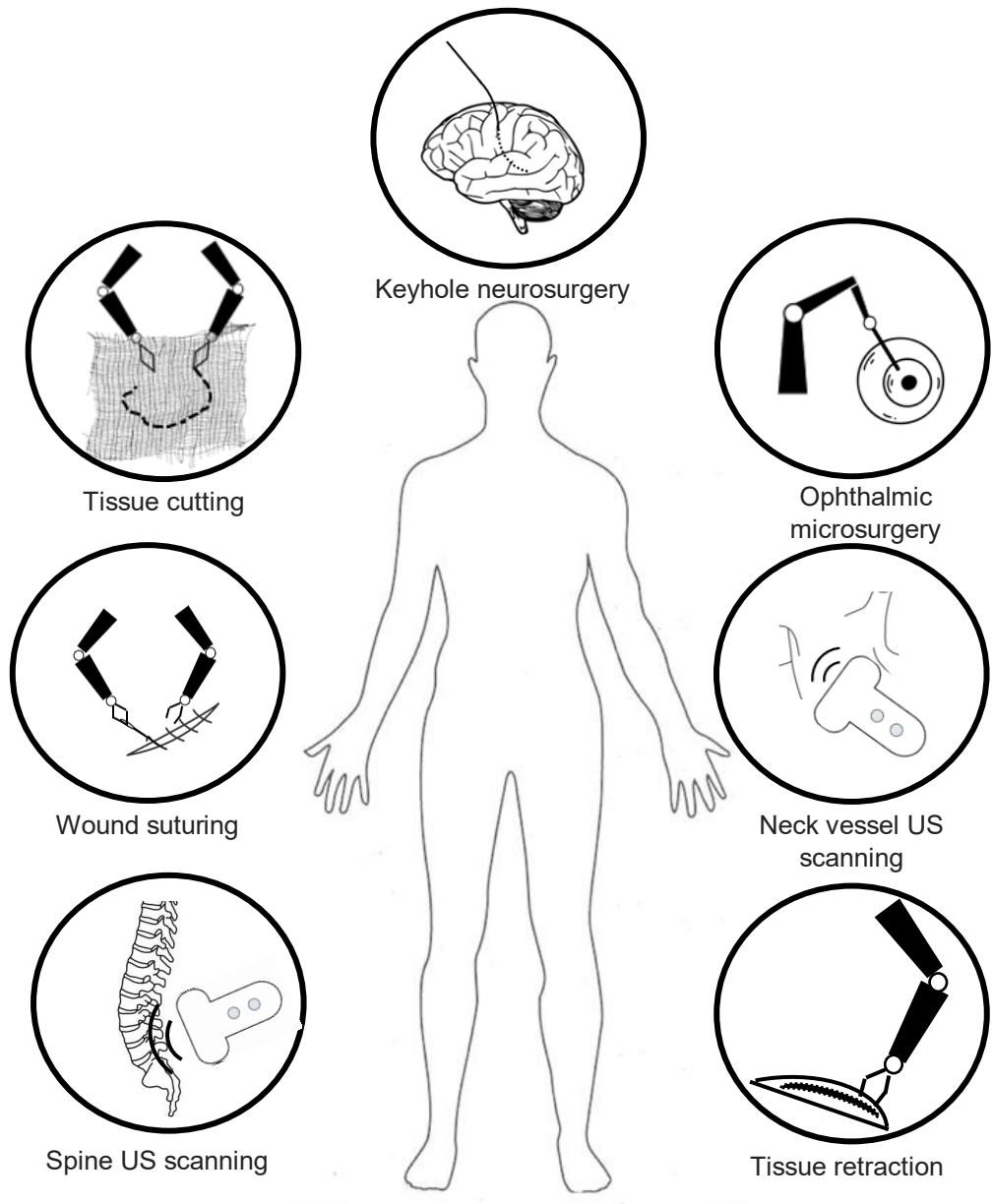

Fig. 1. The seven different robotic surgical tasks contained in this review

To exhaust the published review articles of the concerning fields and extract the most relevant ones, we searched keywords on the database and excluded the irrelevant articles. The articles extraction pipeline is shown in Fig. 2. We also counted the number of studies that applied DRL in medical scenarios in the last 5 years. In Fig. 3, the number of articles on the application of DRL in medical imaging, medical robotics, and dynamic treatment regime in recent 5 years are listed.

We can see that the number of studies combining DRL with different medical fields has quickly emerged in recent years, which indicates a growing trend.

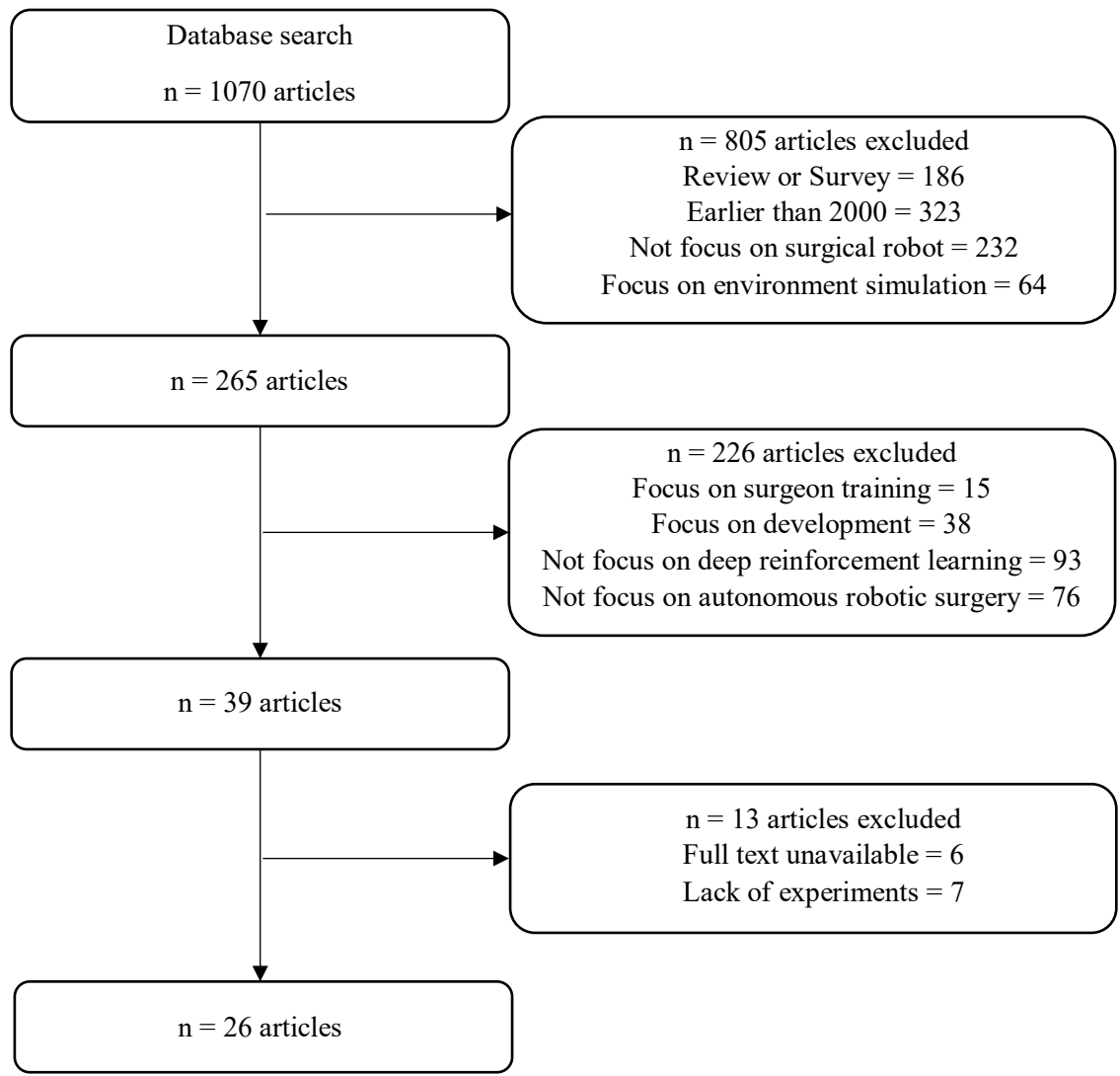

Fig. 2. Articles selection pipeline with keywords "surgical robot", "autonomous", and "deep reinforcement learning" according to PRISMA [81]

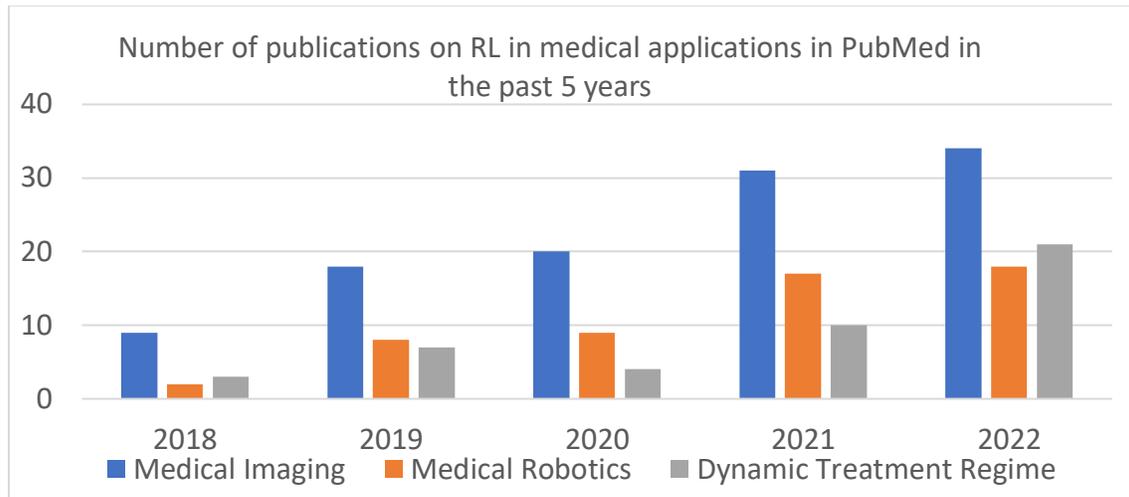

Fig. 3. Statistics of the number of publications on RL in three main medical applications in PubMed in the past 5 years. The combination of DRL with autonomous surgical robots and other medical fields have a rising trend in the last 5 years.

The rest of the chapters are organized as follows: Section 2 briefly introduces the fundamental theories in RL. Sections 3, 4, and 5 discuss the latest work of DRL in the fields of pre-operative scanning, intra-body surgery, and Image-guided autonomous robotic surgery, respectively.

## II. Basics of Reinforcement Learning

Before discussing the state-of-art works of DRL in surgical robotics fields, we will first give a general introduction to the fundamental knowledge in Reinforcement learning (RL). Learning through interaction with the environment is the essence of RL [14], which means an agent learns to take action through rewards and penalties and refines its policy accordingly.

The fundamental of RL includes five essential elements: agent, environment, action, state, and reward. In the context of surgical robots, an example of it can be illustrated in Fig. 4, where the robot (agent) works at the surgical site of a human body (environment), moving the probe to find a feasible scan plane for the sacrum, and obtaining the current position information of probe via real-time ultrasound (US) images. At each time step, the robot possibly gets a positive or negative reward based on the current US image, which guides the robot toward the standard scan plane.

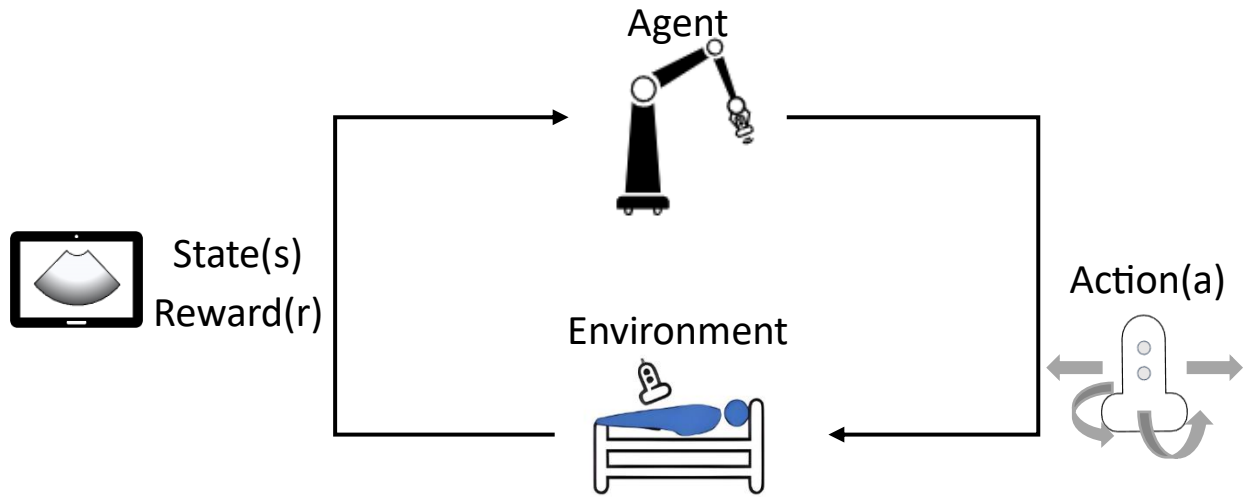

Fig. 4. An illustration of agent-environment interaction in RL under the context of surgical robots

## [Markov Decision Process]

Markov Decision Process (MDP) is always used to formally describe the above-mentioned agent-environment interaction, which consists of [17, 18]:

- State Space ($S$): The set of possible states the agent can be. Each state represents a particular configuration or situation in the environment. In surgical robot scenarios, the state is often chosen as the robot's pose and target.
- Action Space ($\mathcal{A}$): The set of possible actions that the agent can take. Actions are the choices available to the agent in each state. Actions can be discrete or continuous depending on the accuracy requirements. It is preferred to be chosen as continuous in some safety-critical scenarios, which need high accuracy of control, e.g., in intra-body surgery.
- Transition ($\mathcal{T}$): The transition probabilities that describe the dynamics of the environment. They specify the probability of transitioning from one state $s$ to another state $s'$, $if$ the agent takes action $a$, which is represented as $T(s'|s, a)$. It includes important information about the robot-environment interaction, e.g., the interaction between the US probe and tissue.
- Reward (R): The immediate feedback that the agent receives from the environment for its action. It quantifies the desirability or value associated with transitions between states. The reward function is typically denoted as $r = R(s', a, s)$. It is commonly designed to guide the robot to achieve its goal. For example, for standard scan plane navigation in robotic ultrasound scanning, the reward is commonly designed to be the pose improvement of the probe to the target pose.
- Discount factor (γ): A value between 0 and 1 that determines the importance of future rewards compared to immediate rewards. It determines the preference of the agent for immediate rewards or long-term cumulative rewards.

Given $MDP(S, A, T, R, \gamma)$, the agent chooses the action at state $s$ with the observation $o$ it receives according to the policy $\pi(a|s)$. When the policy is deterministic, $\pi$ is a mapping from state $s$ to action $a$; when the policy is stochastic, $\pi$ represents the possibility of selecting action $a$ at state $s$.

The goal of RL is to find an optimal policy $\pi^*$ that maximizes the expectation of cumulative return, which is denoted as:

$$\pi^* = \max_{\pi} \mathbb{E}[\sum_{t=0}^{T} \gamma\, r_t],$$

where $r_t$ is the reward at time $t$, and $T$ is the time horizon.

To be noticed that sometimes the full state information is not available for the agent, but only a part of it, instead. The agent has to predict the state information given an observation. For example, the ultrasound scanning robot has to detect its current position according to the real-time US image. In this case, the process is a partially observable MDP (POMDP) [29]. And the set of the state information that is observable for the agent is called Observation. In this case, the policy $\pi$ is dependent on observation $o$ instead of state $s$.

Besides, in this review, only model-free RL algorithms are focused on. Therefore, the transition is assumed to be unknown.

# III. Deep Reinforcement Learning in Surgical Robotics

In this section, we will highlight the state-of-art studies of DRL for surgical robotics applications and discuss them in three parts: pre-operative scanning, intra-body surgery, and percutaneous surgery. We will focus on how they formulate the problem in a DRL-based framework and different methodologies applied to augment RL to meet some surgery-specific requirements, such as risk analysis.

[Pre-operative Procedure]

Surgical images are obtained using various imaging modalities and ultrasound (US) scanning is the one that has been widely studied in combination with robotics. Over the past two decades, researchers have begun exploring the potential of robotics in applying US scanning. By equipping the robot arm with a probe, the robot can move the probe to perform US scanning on the patient. The accuracy, consistency, skill and maneuverability of robotic manipulators can be used to improve the acquisition and utility of real-time ultrasounds [25]. However, to obtain high-quality ultrasound images, it is crucial to navigate the US probe to the correct scan plane [26] and maintain reasonable and consistent probe-skin contact force [27], as illustrated in Fig. 5. Therefore, standard scan plane localization and contact force control are two main challenges in robotic US and so far, there have been several studies utilizing DRL to address them. In Table 1, the methodologies, metrics, and results used in the 9 reviewed papers in this section are listed.

## a. Standard Scan Plane Navigation

A standard scan plane in ultrasound imaging refers to a recommended imaging plane or view commonly used for a specific anatomical structure or diagnostic purpose. Finding the appropriate scan planes is crucial to obtaining good-quality US images. To enable autonomous robotic ultrasound scanning, the robot should be capable of detecting its own position and finding the way toward the standard scan plane of the specific anatomy with the real-time US image it obtains.

Hase et al. [28] proposed a framework in 2020 to train the robot to autonomously navigate to the standard scan plane of the sacrum with the information of a sequence of history US images with Deep Q-Network (DQN) [7]. The agent is trained on the 2D US images acquired by grid covering and moves with 2-DOF actions, namely moving forward or backward. When moving closer to or further from the desired scan plane, the agent receives positive or negative rewards. A binary classifier that determines whether the robotic probe is at the standard scan plane, depending on the current US image, is used to make the agent stop at the correct position.

One of the limitations in [28] is that the probe is assumed to find the scan plane by moving in a 2D space, which is unavailable in real US scanning scenarios, where the relative pose between the probe and sacrum is not static. Therefore, in Li et al. [31], the state and action space is designed to be the 6D-pose and -twist of the probe so that the learned policy is no longer restricted to the collected data. The agent receives a reward proportional to the pose improvement of the probe at each time step. Besides, the quality of the US image is also considered in this work by evaluating the pixel-wise confidence and giving corresponding rewards to the agent.

An agent that can recognize different anatomies and find the nearest one according to its current position can be advantageous for its flexibility in a real US scanning scenario. In Li et al. [33], an agent is trained to find three different spinal anatomies with a given standard view recognizer for different spinal anatomies.

Real US images as observations given to the agent can be noisy, which makes the agent hard to predict the real state correctly. One of the methods addressing this issue is presented in Bi, et al. [39], which navigates the agent to the scan plane of the neck vessel. The study segments the US images with a pre-trained U-Net [40] in advance and provides the segmented mask to the agent, where the area of interest has pixel values of one, while other areas have values of zero. Compared to real US images, it is much easier for the agent to extract information from segmented masks, which only contain binary values.

Besides, Li, et al. [41] and Milletari, Fausto, Vighnesh Birodkar, and Michal Sofka [42] proposed DRL-based frameworks for training an agent that guides a novice operator to find the standard scan planes in transesophageal echocardiography and chest sonography, respectively.

## b. Pose and Force Control

The way in which the ultrasound probe is positioned and controlled can have a significant impact on both the quality of the resulting ultrasound images and the overall safety of the robotic ultrasound system. It is essential to carefully control the pose and force used when operating the probe, as any errors or inconsistencies can compromise the quality of the imaging and potentially cause harm to the patient or the system itself. A system can ensure imaging quality while minimizing potential risks or complications by taking a deliberate approach to probe control. Unlike rigid objects, force control of the US probe should consider the compliance of the patient body. Besides, errors caused by target movement have to be also compensated. Both of them are hard to be accurately modeled. However, by taking advantage of DRL's model-free and end-to-end properties, the control of the US probe can be solved without explicitly modeling.

In Ning, et al. [42], an agent is trained with Proximal Policy Optimization (PPO) [21] to autonomously control the pose and force of the US probe with a force-to-displacement admittance controller. The agent has to provide proper 2D input command for the controller, namely the desired torque of the US probe in long- and short-horizontal-direction, as illustrated in Fig. 5, to keep the vertical between the probe and the scanned surface with suitable contact force. A 6-D force sensor is attached to the robot end-effector to give the force feedback to the agent. A positive reward is given when the vertical contact force is suitable and the horizontal contact force is small enough, which means the probe is approximately vertical to the scanned surface. A similar work is done in [46], however, with an inverse RL method to study the reward function.

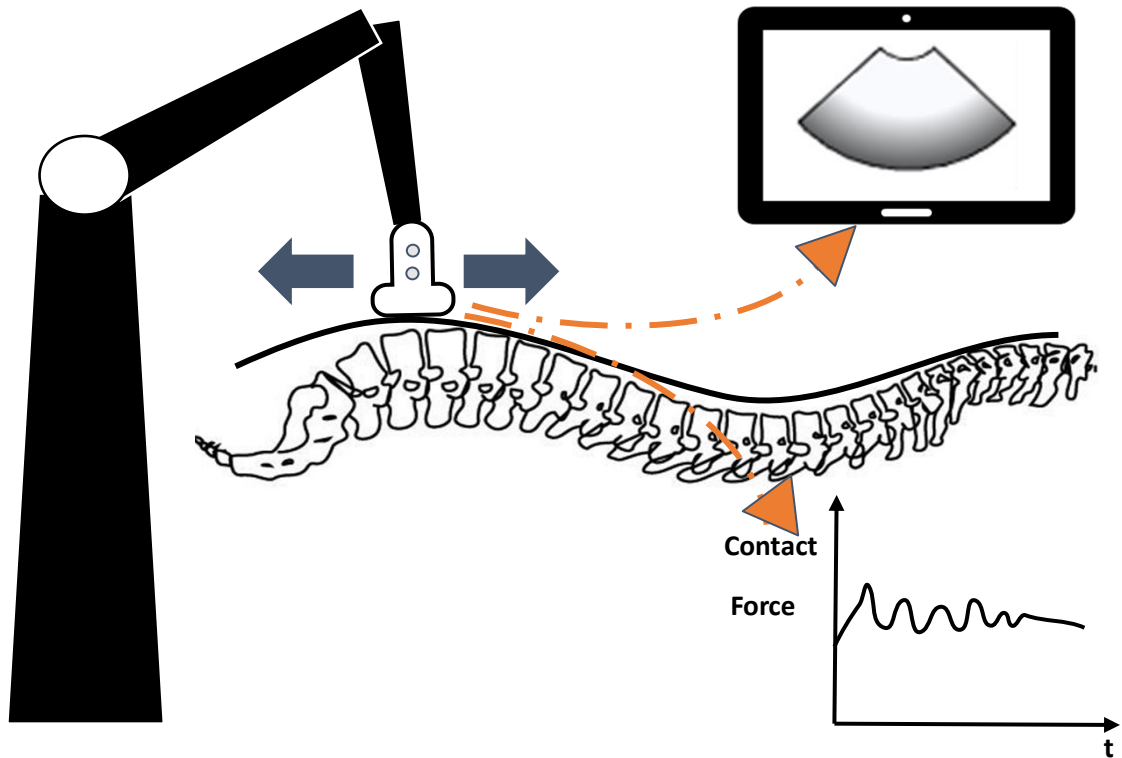

Fig. 5. The robot moves the probe to find the standard scan plane while keeping the contact force in a suitable range.

Differing from [42] and [46], in Ning, et al. [44], the scene image captured by a RGB camera is also provided to the agent as observation. From the scene image, the agent can extract the information of both its own pose and the target pose. In the study [45], a convolutional autoencoder (CAE) and a reward prediction network are employed to achieve two objectives simultaneously. Firstly, the CAE is used to decrease the dimensionality of the observation space, allowing for more efficient data processing. Secondly, the reward prediction network encodes force and ultrasound image information into the scene image, enhancing the resulting image's quality. By utilizing these techniques, the researchers improved the overall efficiency of the system.

### [Intra-body Procedure]

Recently, flexible surgical robotic systems have been developed to improve intra-body surgery in the narrow areas of the human body. However, the teleoperation of surgical robots can be exhausting and needs long-term training time. More and more researchers have been working on the possibility of automating difficult surgical handling tasks, e.g., tissue cutting, suturing, knot tying, and tissue retraction, to reduce the surgeons' workload [64]. However, the large quantity of soft tissue in surgeries, including organs, blood vessels, and muscles, possess inherent compliance and deformability, making their manipulation challenging and requiring modeling and planning with high accuracy and complexity. Therefore, some studies have tried to unleash the model-free property of DRL in automating the tissue manipulation tasks in MIS, including tensioning, suturing and retraction. In Table 2, the methodologies, metrics, and results used in the 8 reviewed papers in this section are listed.

| Ref. | Description | Algorithm | Observation | DOF of Action | Reward | Result |
|---|---|---|---|---|---|---|
| [28] | Navigation towards the standard scan plane of sacrum | DQN | Sequential US images | 1 | + moving closer<br>- moving further | Policy correctness of 79.53% and reachability of 82.91% |
| [31] | Navigation towards the standard scan plane of the spine with consideration of US image quality | DQN | Sequential US images | 6 | + pose improvement or image quality improvement<br>- unallowable pose | 92% and 46% success rate in intra- and inter-patient settings, respectively |
| [33] | Navigation towards different standard scan planes of spinal anatomies | DQN | Sequential US images | 6 | + pose improvement or image quality improvement<br>- unallowable pose | Pose error ~ 5.18mm/5.25° for intra-patient settings; Pose error ~ 2.87mm/17.49° for inter-patient settings. |
| [39] | Navigation towards the standard scan plane of carotid vessels | A2C | Sequential segmented US images + Sequential vessel area changes | 3 | + vessel area improvement<br>- too small vessel area | 91.5% and 80% success rate in simulated and real environment, respectively |
| [41] | Guidance for novice operators in moving TEE probe towards the standard scan plane of heart with pressure awareness | DQN | Sequential US images | 3 | + pose improvement<br>- unallowable pose | Pose error of 2.72 mm / 2.69° and 8.15 mm / 5.58° without and with pressure awareness, respectively |
| [42] | Guidance for novice operators in obtaining correct US images of anatomy of interest | DQN | Sequential US images | 4 | + moving closer<br>- moving further | 86.1% success rate in giving correct guidance |
| [43] | Force control between probe and phantom | PPO | 6-D contact force | 2 | + small horizontal force<br>- big horizontal force or too big or too small vertical force | Difference of skin area in US images within 3 ± 0.4% from the hand-free scanning approach |
| [44] | Force control between probe and phantom | PPO | Single encoded RBG scene image | 3 | + moving closer to the target surface, good US image quality, correct relative position<br>- otherwise | 93% success rate in getting feasible US images |
| [48] | Force control between probe and phantom | PPO + Inverse RL | Contact force and torque, and corresponding linear and angular speed | 6 | Reward shaping via inverse RL | Posture error of 2.3±1.3° and 1.9±1.2° in X and Y axis compared to manual operation |

Table 1. The formulation, methodologies and results of the reviewed papers in the section on pre-operative planning

## a. Tensioning

Robotic surgery has revolutionized the medical field, and an electric knife is an effective tool for cutting and removing thin tissues. However, the electric knife alone may not be enough to cut effectively when it comes to deformable soft tissue. This is because soft tissue needs to be held in tension to be cut most effectively. Therefore, a second tool is required to pinch and tension the material while cutting. This technique is illustrated in Fig. 6, which demonstrates the use of two tools cooperatively to cut soft tissue. The first tool, the knife, cuts the tissue while the second tool, which pinches and tensions the material, helps the knife cut more effectively. This technique is particularly important in robotic surgery, where precision is critical, and using multiple tools can help ensure the surgery is successful. To let the robot autonomously assist the surgeon in cutting. The robot has to learn the tensioning policies for different cutting contours.

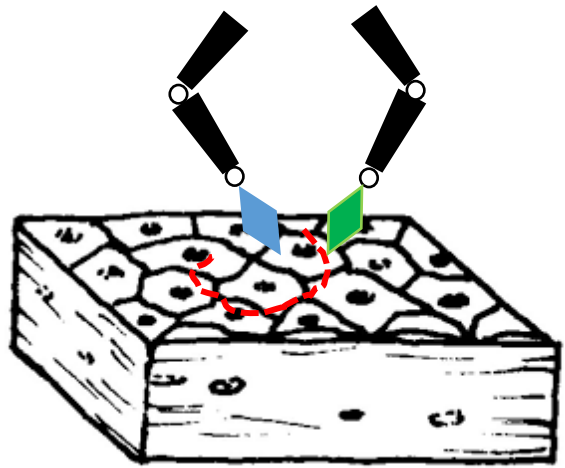

Fig. 6. The blue manipulator pinches the grey point and tensions the tissue, while the green manipulator is responsible for cutting (e.g., in Endoscopic Submucosal Dissection, ESD)

In Thananjeyan, et al. [50], a finite-element model is first developed for simulating the deformation and cutting of tissue. Then, considering the kinematics constraints of the surgical robot arm, the cutting outline is divided into several subdivision segments in advance. The agent is lastly trained in Trust region policy optimization (TRPO) [22] to learn the optimal tensioning policy to minimize the cutting error with a single fixed pinch point. The agent receives sparse rewards at the end of each episode according to the final cutting error.

There needs to be more than a single tensioning point to assist cutting, when the cutting pattern is complex, for example, the cutting contours are zigzag and have to be divided into many segments. Therefore, in [51, 52], an improved pipeline is proposed to address this limitation. Specifically, a pinch point is chosen for each cutting segment instead of the whole contour and the agent learns different tensioning policies for each pinch point in a similar way as in [50]. Compared to [50], the improved method shows more accurate and robust performance, when handling complex cutting contours.

## b. Suturing

Suturing is a critical step in wound closure during surgeries and in robot-assisted surgeries. However, robotic suturing can be laborious for novice operators. A collaborative robot that autonomously assists surgeons in performing some sub-tasks in robotic surgeries can effectively operators' fatigue. So far, utilizing DRL on surgical collaborative robots can learn how to autonomously collaborate with surgeons in teleoperated suturing process, as illustrated in Fig. 7. In Table 3, the methodologies, metrics, and results used in the 9 reviewed papers in this section are listed.

In Varier, et al. [53], an agent is trained to use an assistive Patient Side Manipulator (PSM) to pull the needle, translate it to the next suture point and hand it to the surgeon after the needle is inserted through the tissue with the main PSM by the operator, as illustrated in Fig. 7. To address varying suture styles of users, the users are instructed to perform a running suture without a collaborative robot. The trajectory of a single hand-off task is collected and an algorithm is designed to generate sparse rewards on the trajectory. Then, the agent imitates the operator's hand-off trajectory by maximizing the cumulative collected reward.

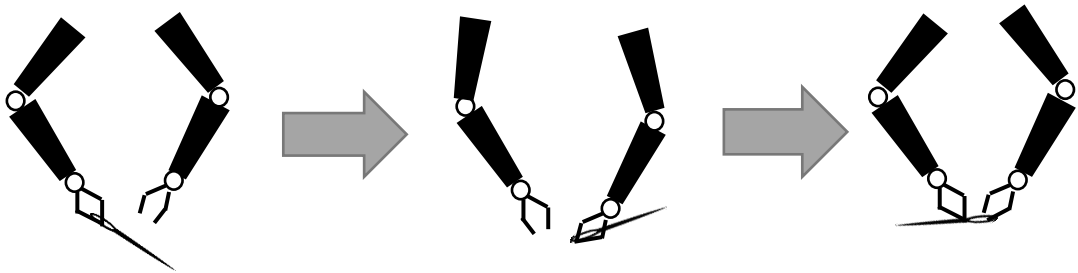

Fig. 7. The main manipulator (left) is operated by the surgeon, which inserts the needle through the tissue, while the assistive manipulator (right) pulls the needle out and hands it to the main manipulator.

However, in [53], the state of the agent is respect to a fixed frame, which makes the learned policy strongly depend on the selection of the frame. In Chiu, et al. [55], an improved method is proposed to address this issue by designing action spaces as being respect to the ego-centric frame, which means the policy depends on the relative position and orientation of the assistive PSM relative to the main PSM and therefore can be directly applied to different robot configurations.

c. Retraction

Another kind of tissue manipulation task in robotic surgery is tissue retraction. I.e., to uncover the underlying anatomical region, the tissue is repeatedly held and pulled back in MIS [57]. To autonomously perform the tissue retraction task, the robot has to find the position of the tissue, move closer to it and grasp it to the target position.

In Pore, et al. [58], a robot learns to approach the tumor from its initial position and retract it to the target position from human demonstrations. The position of the tumor is assumed to be known from the pre-operative data. The agent gets the reward based on whether it moves closer to the tumor or target position before or after grasping. Human demonstrations are collected to enable imitation learning. The agent is trained with Generative adversarial imitation learning (GAIL) and PPO, where PPO acts as the action generator.

Safety is always the priority in surgeries, especially in tissue retraction, where the robot directly interacts with the tissue. However, due to the model-free property of DRL, the safety of the learned policy is always hard to verify, which leads to significant potential risks in surgery. In another work by Pore, et al. [60], a framework for robotic tissue retraction incorporates the safety constraints during the DRL training with formal verification, which adds a penalty term in the reward function for unsafe actions and evaluates the safety of the learned policy [61]. The proposed method shows a large reduction in the safety violation rate, compared to [58].

In [58] and [60], the agents are assumed to have access to the full-state information, e.g., the robot joint angles and tissue position, which makes the policy largely depend on the accuracy of state extraction and lack robustness against the patient movements. In Scheikl, et al., [62], a vision-based framework for robotic tissue retraction is proposed. The agent is trained with the simulated RGB scene image and a translation model is trained to translate the observation function in simulation to the one in reality using domain adaptation. The trained agent achieves a success rate of 50% in real surgical scenarios.

## [Percutaneous Procedure]

Percutaneous techniques are increasingly used in many surgical scenarios, including neurosurgery and ophthalmic surgery. The practical advantages include lower complexity rates and faster recovery time. It involves the precise insertion of a thin, hollow needle into specific anatomical structures for diagnostic, therapeutic, or monitoring purposes. This technique allows surgeons to access internal body tissues, organs, or vessels without the need for open surgery, minimizing patient discomfort, reducing the risk of complications, and promoting faster recovery.

### a. Needle Insertion

Needle insertion surgery is common in various medical fields, including keyhole neurosurgery [65] and ophthalmic surgery [66]. Conventionally, a rigid needle is commonly used in these procedures. There has been some research on DRL-based needle path planning. In Keller et al. [73], an agent is trained to control the yaw, pitch, and depth of the needle to achieve the goal position in ophthalmic surgery with optical coherence tomography (OCT) image as observation. In Gao, et al. [74], an agent is trained to provide a remote center of motion (RCM) [75] recommendation in brain surgery. The author considers three aspects to evaluate the quality of RCM, namely clinical obstacle avoidance (COA), mechanically inverse kinematics (MIK) and mechanically less motion (MLM) and the reward function is also designed based on these three aspects.

However, it can be challenging for rigid needles to find safe trajectories to insert toward the target without touching some critical anatomies, e.g., blood vessels, especially when the structure is complicated. Therefore, steerable needles have attracted much attention in the last decade due to their flexibility. Accurate path planning is one of the most crucial factors for successful steerable needle insertion, where the tissue-needle interaction has to be considered. In Lee, et al. [68], an agent is trained to perform pre-operative path planning for steerable needles in keyhole neurosurgery with DQN. The environment is simulated by segmenting 2D MRI images into obstacles and obstacle-free areas. The agent controls the bevel direction rotation and insertion depth to insert the needle toward the target area. The agent is rewarded when achieving the goal and punished when entering an unsafe area, e.g., a blood vessel. In Kumar, et al. [69] and Segato, et al. [72, 79], similar frameworks are proposed, however, with 3D MRI images to enable 3D path planning and continuous actions.

Pre-operative path planning provides initial planning, including the insertion point and a rough global plan. However, due to unexpected anatomical movements or needle-tissue interactions, the pre-planned path can be violated and therefore, intra-operative replanning is needed. Furthermore, it is also essential for the surgeon to easily detect the risk potential (possibility of the needle entering unsafe areas) of the re-planned path. In Tan, et al. [70], a universal distributional Q-learning (UDQL) [71] based training framework is proposed to enable fast replanning and risk management. In UDQL, the expected Q-value is parameterized with a value distribution, so that only a distribution with a high mean Q-value and low variance can be considered a safe plan.

| Ref. | Description | Algorithm | Observation | DOF of Action | Reward | Result |
|---|---|---|---|---|---|---|
| [50] | Tensioning policy for the selected pinch point | TRPO | Cutting trajectory and fiducial points locations | 2 | Sparse reward according to the final cutting error, when episode ends | Improvement of 43.3% compared to non-tension baseline in term of cutting error |
| [51, 52] | An improved pipeline enabling selecting multiple pinch points for different cutting segments | TRPO | Cutting trajectory and fiducial points locations | 2 | Sparse reward according to the final cutting error, when episode ends | Improvement of 50.6% compared to non-tension baseline in term of cutting error |
| [53] | Autonomous collaborative needle hand-off task of PSM | Q-Learning | 3-D Position of robotic tip | 3 | Sparse reward according to the data points on user-defined trajectory | Dissimilarity between learned trajectory and reference trajectory with mean and standard deviation of [2.857mm, 1.488mm, 0.774mm] and [3.388mm, 2.286mm, 1.808mm] |
| [55] | Autonomous collaborative needle hand-off task of PSM in ego-centric spaces | DDPG + BC | Relative position and quaternion | 6 | + reaching target pose<br>- collision | 97% and 73.3% success rate in simulation and real-world environment, respectively |
| [58] | Robotic tissue retraction learning from expert demonstration | PPO + GAIL | gripper state, end-effector location, | 3 | + moving closer to tumor or target position<br>+ moving further from tumor or target position | Average tumor exposure percentage of 84% and 90% in simulation and real-world environment, respectively |

| | | | target location | | | |
|---|---|---|---|---|---|---|
| [60] | Robotic tissue retraction considering safety properties | PPO + Formal Verification | Gripper state, end-effector location, target location | 3 | + moving closer to tumor or target position<br><br>+ moving further from tumor or target position or violating safety constraints | Safety violation rate of 3.07% and violation rate reduction of 24%, compared to non-safety method |
| [62] | Robotic tissue retraction with sim-to-real | PPO + DCL | Sequential translated scene image | 2 | + moving closer to tumor or target position<br><br>+ moving further from tumor or target position | 50% success rate in real world environment with raw camera images as input |

Table 2. The formulation, methodologies and results of the reviewed papers in the section of intra-body surgery.

## b. Catheterization

Catheterization is one of the most commonly used procedures in endovascular intervention. The catheter is guided to the target of the disease in the vasculature along with treatments such as stenting, embolization, and ablation, as illustrated in Fig. T8. However, the guidance of the catheter is not trivial. The surgeon needs to manipulate the catheter with limited 2D fluoroscopic information and minimize unwanted excessive tissue contacts. Due to the difficulty of manually operating the catheter, robot-assisted catheterization has been researched in the last decade and DRL is one of the promising methods for its path or trajectory planning.

In Chi, et al. [76], an agent is trained to optimize the catheterization trajectory demonstrated by the experts. The expert demonstrations are first parameterized with dynamic motion primitives (DMP). The agent adjusts the parameters of DMP to optimize the trajectory, guiding the catheter tip towards the desired vessel plane while matching the trajectory with the vessel centerline as much as possible. The agent is trained with Path Integral ($PI^2$) [77] algorithm, which is a robust RL implementation based on trajectory rollouts. The environment is based on vascular models with flow simulation. Furthermore, Chi, et al. [78] proposed a closed-loop catheter control framework based on GAIL to imitate the expert catheterization demonstration. An electromagnetic (EM) tracking sensor is attached to the catheter tip to take into account its real-time pose to enable intra-operative control. The agent is trained to imitate the five-motion primitive of the expert's hand demonstration, namely pull, push, clockwise rotation, anti-clockwise rotation, and stand-by. Besides, in Omisore, et al. [80], DRL is utilized to tune the parameters of a PID controller in real time, to let it adapt to different blood flow settings.

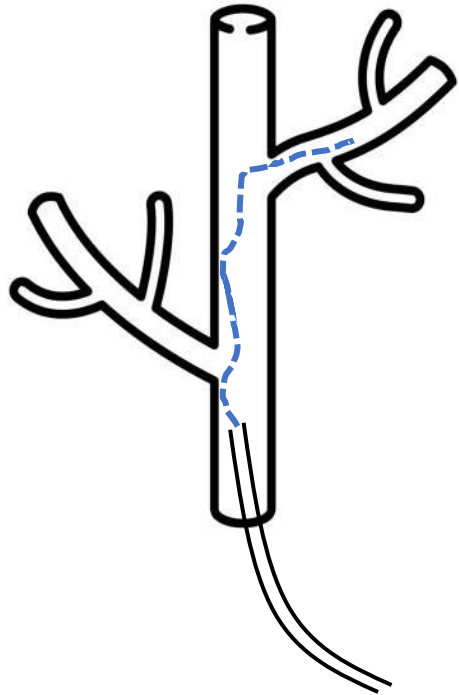

Fig. 9. The catheter is guided to the target position in vessel. The blue dot line is the planned path.

| Ref. | Description | Algorithm | Observation | DOF of Action | Reward | Result |
|---|---|---|---|---|---|---|
| [73] | Needle path planning for Ophthalmic microsurgery | DDPGfD | Height maps of two corneal surfaces | 3 | Sparse reward when reaching the target position | Perforation-free percent depth of 84.75% ± 4.91% |
| [74] | RCM recommendation for keyhole neuro surgery | PPO | Target position, Robot joints position | 2 | Sparse reward according to positioning accuracy, solvable inverse kinematics and mechanical joint motion | 93% success rate of finding optimal RCM |
| [68] | Navigation of steerable needle towards the target position in brain in 2D space | DQN | 2D map based on segmented MRI images | 2 | + position improvement | 93.6% success rate of achieving target position |
| [69] | Navigation of steerable needle towards the target position in brain in 3D space | DDPG | 3D map based on segmented MRI images | 2 | + achieve goal or in safe area <br> - in unsafe area | Outperforms RRT* under different quantiles of constraints |
| [72] | Navigation of steerable needle towards the target position in brain in 3D space | GAIL | 3D map based on segmented MRI images | 6 | + achieve goal <br> - obstacle collision | Average targeting error of 1.34 ± 0.52 mm in position and 3.16 ± 1.06 degrees in orientation |
| [79] | Navigation of steerable needle towards the target position in brain in 3D space | GA3C | Sequential frames of 3D map based on segmented MRI images | 6 | + achieve goal <br> - obstacle collision | Outperforms RRT* under different quantiles of constraints |

| [76] | Optimization of catheterization trajectory obtained from demonstration | $PI^2$ | Catheter tip pose, Target position | 2 | + position improvement, vessel centerline alignment | Average targeting error of 2.92mm and path length of 258.67mm |
|---|---|---|---|---|---|---|
| [78] | Navigation of catheter tip to the target position | GAIL+ PPO | Catheter tip pose, Target position | 3 | + position improvement<br>- obstacle collision | 82.4% success rate of achieving target position |
| [80] | Adaptation of PID controller parameters for catheterization | DQN | Catheter tip pose | 3 | + position improvement | Average error of 0.003 ± 0.0058 mm with respect to setting point |

Table 3. The formulation, methodologies and results of the reviewed papers in the section of percutaneous surgery.

# [Conclusion]

This literature review has provided an overview of the application of Deep Reinforcement Learning (DRL) in surgical robots. We divided the state-of-art works that applied DRL on surgical robots into three main fields: skin-interfaced, intra-body, and percutaneous, discuss how they formulate the problem based on RL-framework, and compare their methodologies and limitations. Based on the outstanding performance of DRL in these works, the integration of DRL algorithms into surgical robotic systems has the potential to revolutionize the field of robotic-assisted surgery by enhancing the autonomy and decision-making capabilities of these systems.

The technology of DRL is in its youth and still suffers from some limitations, e.g., low data efficiency, expensive to train in the real world, and lack of safety guarantee. Looking forward, further research is needed to refine and optimize DRL algorithms for surgical applications. This includes the following points: Firstly, more efficient training methodologies. Currently, most DRL algorithms are very sample inefficient compared to other deep learning methods. Secondly, a more accurate simulation environment. In surgical scenarios, there exist a lot of deformable bodies interaction, which are much harder to simulate, compared to rigid bodies interaction. Thirdly, addressing safety concerns. Safety is always the priority in surgeries. This could include risk analysis or interpretability of the model. Lastly, conducting clinical trials to evaluate the effectiveness and reliability of DRL-based surgical robots because, so far, few DRL-based robots have been tested in real clinical scenarios.